\documentclass[12pt]{article}
\usepackage[a4paper]{geometry}
\usepackage{fullpage}
\usepackage{graphicx}
\usepackage{arydshln}
\usepackage{epsfig}
\usepackage{enumerate}
\usepackage{caption}
\usepackage{subcaption}
\usepackage{wrapfig}
\usepackage{float}
\usepackage[english]{babel}
\usepackage{amsmath}
\usepackage[nottoc]{tocbibind}
\usepackage{xcolor}
\usepackage{sectsty}
\usepackage{titling}
\usepackage{tfrupee}
\usepackage[T1]{fontenc}
\usepackage{titlesec}
\usepackage{blindtext}
\usepackage{parskip}
\usepackage{makeidx}
\usepackage{multirow}
\usepackage{etoolbox}
\usepackage{url}
\makeatletter
\providecommand{\subtitle}[1]{
  \apptocmd{\@title}{\par {\large #1 \par}}{}{}
}
\makeatother
\providecommand{\keywords}[1]{\textbf{\textit{Keywords---}} #1}
\title{Unsupervised Change Detection in Hyperspectral Images using Feature Fusion Deep Convolutional Autoencoders}
\author{Debasrita Chakraborty, Ashish Ghosh}
\date{}
\begin{document}
\maketitle

\begin{abstract}
Binary change detection in bi-temporal co-registered hyperspectral images is a challenging task due to the large number of spectral bands present in the data. Researchers, therefore, try to handle it by reducing dimensions. The proposed work aims to build a novel feature extraction system using a feature fusion deep convolutional autoencoder for detecting changes between a pair of such bi-temporal co-registered hyperspectral images. The feature fusion considers features across successive levels and multiple receptive fields and therefore adds a competitive edge over the existing feature extraction methods. The change detection technique described is completely unsupervised and is much more elegant than other supervised or semi-supervised methods which require some amount of label information. Different methods have been applied on the extracted features to find the changes in the two images and it is found that the proposed method clearly outperformed the state of the art methods in unsupervised change detection for all the datasets.
\end{abstract}
\section{Introduction}
A remotely sensed hyperspectral image (HSI) \cite{jin2018intrinsic} is a complex image data which has multiple bands (narrowly spaced). Each band of a remotely sensed hyperspectral image of a particular region captures multiple wavelengths of the electromagnetic spectrum. Hyperspectral imaging seeks to utilize the spectral information which has more information about the region than multi-band images. Hyperspectral images are basically, image cubes that consist of hundreds of spatial images. Each spatial image, or spectral band, records the responses of ground objects at a different wavelength. For example, the NASA AVIRIS (Airborne Visible/Infrared Imaging Spectrometer) captures the spectral images in 224 contiguous spectral bands at the visible and near-infrared regions. The spectral range in hyperspectral data extends beyond the visible range (ultraviolet, infrared).  All these images may not be visible to the human eye, but contain important information to identify ground objects. Hyperspectral images helps us to explore the unexplored, by visualizing information invisible to human eyes.

However, since it contains multiple bands, sometimes the information about different changes is often found in different bands. That is why, it becomes a complicated task to detect changes between two registered hyperspectral images of the same place when there are diverse minute and bigger changes. To do the same in an unsupervised manner is even more challenging. Researchers have proposed a wide range of such digital change detection methods ranging from supervised,semi supervised and unsupervised methods.

Supervised methods \cite{volpi2013supervised} include training the system with some training labels from the ground truth data. A majority of the pixels (usually 70 percent) are labelled as changed or unchanged. The classifier system tries to learn the changed and unchanged classes through the labels provided. It then detects which of the remaining pixels are changed or unchanged. This kind of supervised change detection may be regarded as a problem of classifying imbalanced classes (as the number of unchanged pixels are more than number of changed pixels). Semi-supervised methods \cite{wu2017semi, yuan2015semi} provide limited training data to the classifier system. There are only a few labelled samples (mostly from the unchanged class). Researchers have also explored different selective labelling techniques like active learning which share resemblance to the semi-supervised methods. Manual labelling of pixels are quite cumbersome and expensive. Moreover for HSIs, the labelling becomes challenging due to the sheer number of spectral bands. Since unsupervised methods \cite{liu2014hierarchical}, mainly rely on detecting the points of dissimilarities between the two images, they do not require any manual labelling. Due to this reason, unsupervised methods are mostly less accurate than the supervised or semi-supervised methods but are much more sophisticated in application.

In this article, an unsupervised change detection system is proposed that makes use of feature extraction by a novel feature fusion convolutional autoencoders (FFCAE). The traditional convolutional autoencoders is modified to suit the feature extraction process for change detection in hyperspectral images. The proposed feature extraction technique addresses three main concerns that arise during change detection in bi-temporal co-registered HSIs:

\begin{enumerate}[(i)]
\item Given two HSI, lower their dimensionality in the most efficient way possible.
\item Detection of minute changes with lesser misses and false alarms.
\item No manual labelling required.
\end{enumerate}

Keeping the problem statement in mind, the proposed solution presents a lower dimensional feature map which is enough to detect the changes between the two co-registered bi-temporal hyperspectral images. The feature extracting FFCAE network has multiple hidden layers in the encoder as well as the decoder part unlike a stacked version of autoencoder and is different from as standard deep convolutional autoencoder (DCAE) as it preserves both the spatial as well as the spectral information in its hidden code layer along successive levels and multiple receptive fields. Since the proposed method makes a fusion of features of lower and higher level features along multiple receptive fields, it is termed as a feature fusion convolutional autoencoder. Results show that the proposed method outperforms all the other methods it has been compared with.

The subsequent section \ref{LS} describe the literature survey on the present topic. Section \ref{PM} provides a detailed overview of the proposed methodology and section \ref{Results} reports the results obtained. Finally, section \ref{conclusion} concludes the manuscript and also suggests the directions of future works.

\section{Literature Survey}\label{LS}
Most change detection algorithms for HSIs \cite{wu2017semi, du2019unsupervised, wang2018getnet, zhan2017change} can detect major changes i.e. major land transformations but they are not able to detect subtle changes, which appear only in few spectral bands like water content, crop growth differences, land cover transitions etc. Noise generated by both external sources (atmospheric effects due to absorption and scattering) and internal sources (instrument noise) also affect the hyperspectral sensors. For this reason, need of robust methods of change detection is necessary.

Supervised models classify the pixels of each single image from different time periods using supervised classifiers and then compare the generated maps to detect changes. Recent supervised methods which show promising performance are using convolutional neural networks \cite{wang2018getnet, zhan2017change, zhao2021spectral} for classification of pixels, but this requires the ground truth. Manually labelling ground reference data is expensive and unfavourably low. This is why, semi-supervised \cite{wu2017semi, de2013semi, roy2013neural} or unsupervised \cite{ghosh2007context} methods of change detection are favoured which do not require extensive efforts of human annotation. A recent semi-supervised method of change detection in hyperspectral images \cite{wu2017semi} involves the use of pseudo-labels. However, such methods are again dependent on human labelling of unchanged pixels. To avoid the expenses of gathering ground truth data, unsupervised change detection methods are much more gracious. This manuscript, therefore primarily concentrates with the domain of unsupervised change detection only.

\begin{figure}
    \centering
    \includegraphics[width=0.5\linewidth]{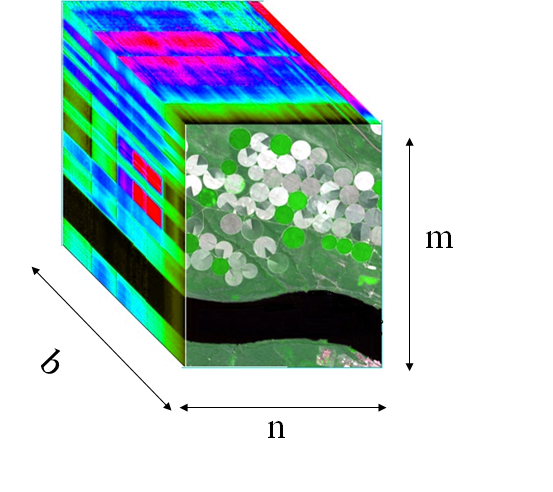}
    \caption{A conceptual diagram of a hyperspectral image.}
    \label{fig:HSI}
\end{figure}

Traditional unsupervised solutions include finding the difference image (DI) and then thresholding it or clustering the pixels. Let a hyperspectral image $\textbf{I}$ be of size $m \times n$ and has $b$ number of bands as shown in Figure \ref{fig:HSI}.
So, each pixel of image $\textbf{I}$, denoted as $\textbf{p}_{ij}$ will be a vector of $b$ dimensions.
\begin{equation}
\textbf{p}_{ij}=\{p_{ij}^1,p_{ij}^2,p_{ij}^3,... , p_{ij}^b\},
\end{equation}
where, $i=1, 2, 3, ..., m$ and $j=1, 2, 3, ..., n$.

Let us consider a pair of bi-temporal co-registered hyperspectral images (with $b$ bands) $I_1$ and $I_2$ of the same location photographed over different times. The assumption of them being co-registered fulfil that for all the pixels in the two images, pixel $\textbf{p}_{ij}^1$ for $I_1$ and pixel $\textbf{p}_{ij}^2$ for $I_2$ represent the exact same point of the original location. The change occurred over the location can be seen by generating a binary change map $C$ such that the $ij^{th}$ pixel $c_{ij}$ of $C$ is marked as changed (by setting the value to one) if the location captured by the $ij^{th}$ pixels for both images has changed and is marked as unchanged (by setting the value to zero) if the location has not changed. The generation of change map may be formally constructed by the equation,
\begin{equation}
C=g(f(I_1) \ominus f(I_2)),
\end{equation}
where, $f()$ is the feature transformation function, $g()$ is the decision function and $\ominus$ is the difference operator.

Without feature transformation, one may often end up struggling with noise which are falsely detected as changes \cite{liu2014hierarchical}. Feature transformation model the relationship between the image object and its real-world geographical feature as closely as possible by reducing the amount of such false alarms and missed changes. Considering only the spectral pixel level information often results in high false alarms \cite{liu2014hierarchical}. For this reason, many methods incorporate pixel level information as well as spatial-context information. Researchers have proposed many methods to handle this problem. In the line of linear feature transformation methods, a PCA based technique is proposed \cite{celik2009unsupervised}, which employs an advanced technique by doing a window based PCA technique for hyperspectral images. This incorporates both the spatial as well as spectral information. However, a linear transformation is often not enough to represent the complex relationships between the two images. So, researchers have proposed non-linear methods like  iteratively re-weighted multivariate alteration detection (IRMAD) \cite{nielsen2007regularized} which has been claimed to be better at dimensionality reduction of hyperspectral images for change detection. This method has recently been supplemented with synthetic fusion of images \cite{wang2015application} called SFI-IRMAD which has shown to give a boost to the change detection performance. Another recent method showed promising results for change detection in hyperspectral images using synthetic fusion of images \cite{han2017unsupervised}. In this case, they have used a custom difference operator that is mainly based on the spectral angle mapper (SAM) \cite{carvalho2011new}. With the popularity of deep learning, some deep convolutional networks have also been proposed \cite{du2019unsupervised, zhao2021spectral}.

A method called S3DCAE \cite{zhao2021spectral} uses a three dimensional deep convolutional autoencoder to extract relevant features from the hyperspectral images. However, the results reported make use of limited labelled samples and falls under the domain of semi-supervised change detection. They claim that their method of feature extraction is robust and thus the extracted features may easily be used for unsupervised change detection as well. Another unsupervised deep convolutional autoencoder based method called DSFANet \cite{du2019unsupervised} is proposed which showed good change detection capabilities. This method uses deep feature extraction of the two images and then does a slow feature analysis on the absolute difference image generated by the two features.

It is quite obvious that all of these methods are basically reducing the dimensions of the two hyperspectral images and have mostly used k-means as the final decision function $g()$ on the difference image generated. All these methods somehow try to reduce the missed changes and false changes as much as possible. Any method which provides a better balance will be deemed superior. Thus, the present manuscript proposes a novel unsupervised method of feature extraction that is capable of detecting changes at a better performance than the above state of the art methods. The details of the proposed architecture is provided in the next section.

\section{Proposed Methodology}\label{PM}
The proposed technology utilises an unsupervised neural network to train on a pair of bi-temporal co-registered \cite{tommaselli2019refining} images of a particular region. The main motivation is that, the neural network learns to compress all the band information into a transformed image of lower dimension, which would retain maximum possible information from all. We have taken four pairs of hyperspectral images for the study. The description of the datasets is provided in the next section.
\subsection{Dataset Description}
Each of the dataset taken in the study are a pair of bi-temporal hyperspectral co-registered \cite{tommaselli2019refining} images. The first pair of images, named as USA dataset, correspond to an irrigated agricultural land in the Hermiston city of USA. The two hyperspectral images, each with a size of 307 $\times$ 241 pixels, were acquired by the Hyperion sensors respectively on May 1, 2004, and May 8, 2007 \cite{hasanlou2018hyperspectral}. The second pair of images are, named as the China dataset, correspond to a farmland in the Jiangsu province of China. These images were acquired by the Hyperion sensors May 3, 2006, and April 23, 2007, respectively \cite{hasanlou2018hyperspectral} and have a size of 420 $\times$ 140 pixels. Both the USA dataset and the China dataset images consist of 154 spectral bands. The third dataset in consideration, named as River dataset, is of the Yangtze river in the Jiangsu Province of China \cite{wang2018getnet}. The two images were captured by the Hyperion sensors on  May 3, 2013, and December 31, 2013, respectively. The River dataset has a size of 463 $\times$ 241 pixels and has 198 spectral bands. It is to be noted that although the intrinsic number of bands for the Hyperion sensors is 242, the above datasets available freely had lesser number of bands as they were already subjected to the band removal process and only the bands with high signal to noise ratio were available publicly \cite{wang2018getnet}. We have not used any kind of pre-processing on the available images and have considered them in our study in the same form as they were available. The Hermiston dataset \cite{javierlopezfandino2018stacked} was aquired by the Hyperion Sensors in the years 2004 and 2007 respectively over the Hermiston City area in Oregon. The image sizes are 390 $\times$ 200 each and includes 242 spectral bands. Out of these 242 bands, only 198 channels were informative. The rest of the 44 channels had the same value in all the pixels and hence had zero entropy (i.e. no information). So, the 44 channels with zero-entropy were omitted and only the 198 channels were considered for further processing. Although, the ground truth showed 5 types of changes related with crop transitions, we converted this to a binary change vs no-change problem and all the 5 change classes were merged to the single change class.


The next subsection describes the particular type of autoencoder configuration that has been used.
\subsection{Proposed Feature Fusion Convolutional Autoencoder (FFCAE)}\label{FFCAE}
\begin{figure}[htp]
    \centering
    \includegraphics[width=1\linewidth]{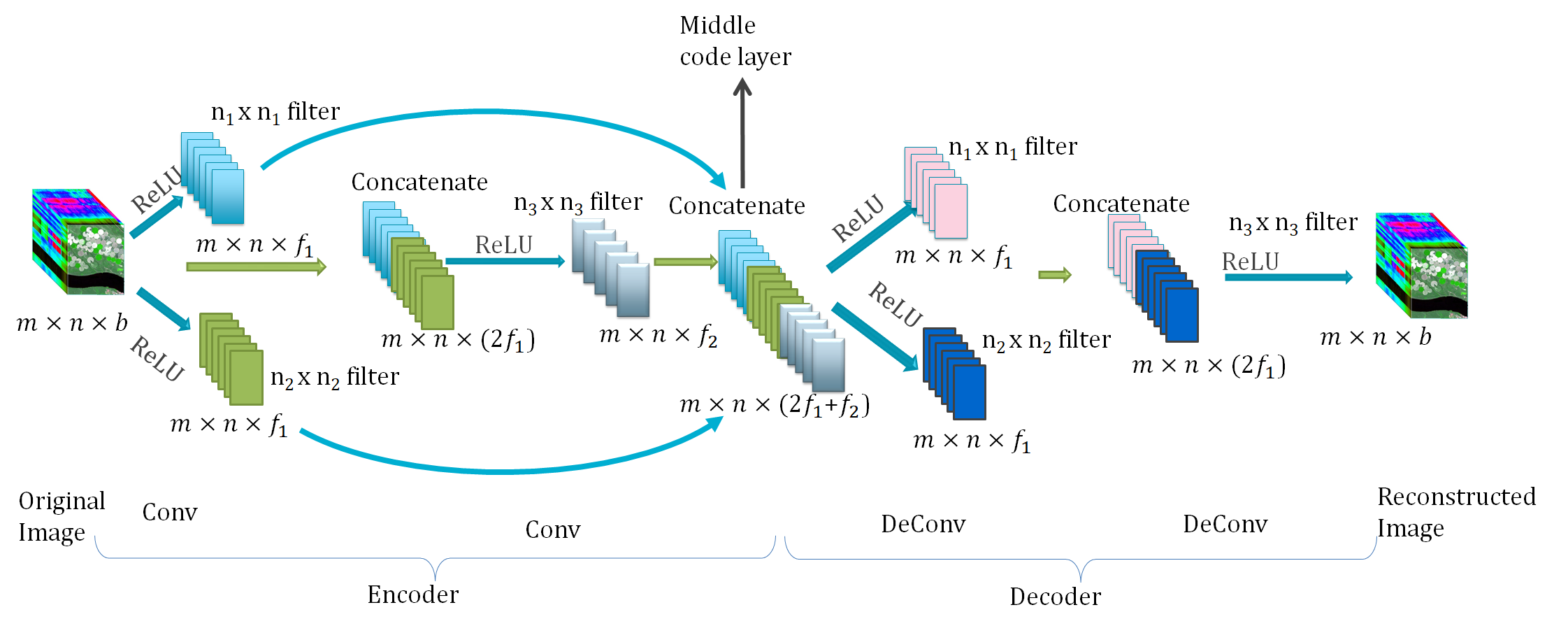}
    \caption{Architecture of the proposed FFCAE (blue arrows represent the trainable weights).}
    \label{fig:FFCAE}
\end{figure}
\begin{figure}[htp]
    \centering
    \includegraphics[width=1\linewidth]{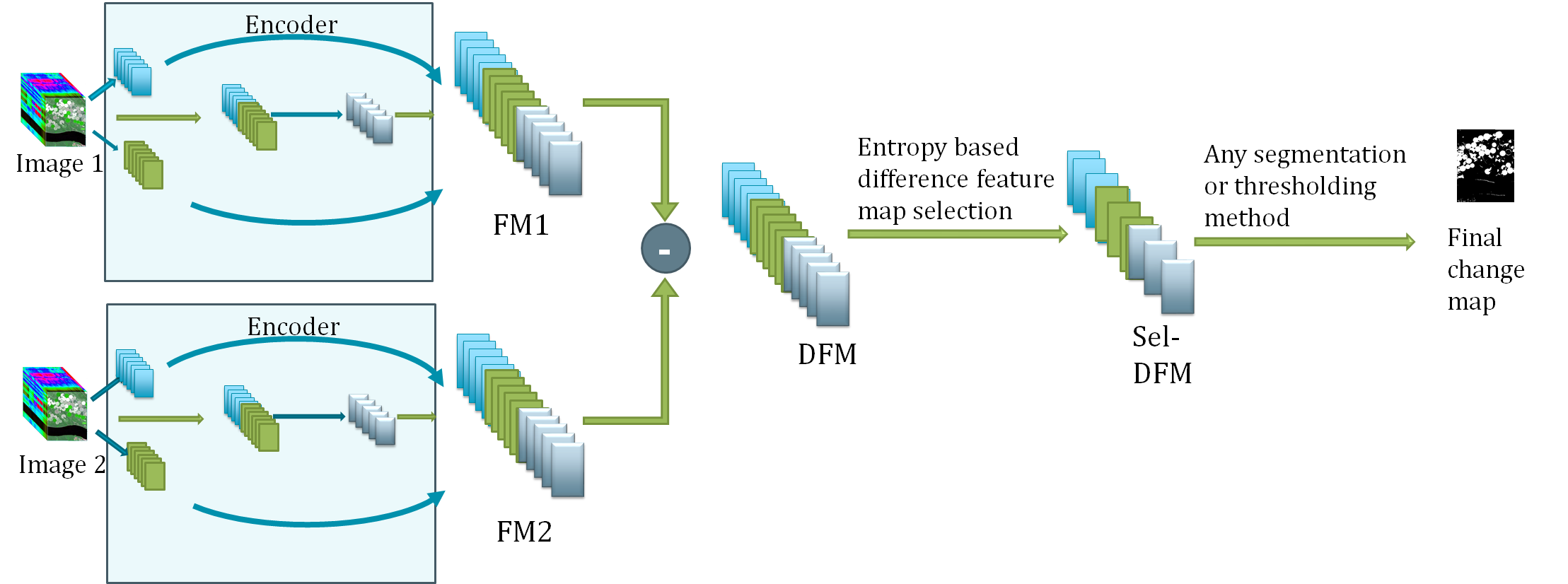}
    \caption{Overall procedure of change detection.}
    \label{fig:overall}
\end{figure}

The main challenge with the afore-mentioned hyperspectral images are that they have a huge number of bands, each of which contain a great deal of information about the topology of the region. Moreover, the dimensions of these images are usually of the order of hundreds on thousands. So, to use a supervised convolutional network (CNN) would become very time consuming. On the other hand, autoencoders would be quite efficient in reducing the dimensions (both in bands as well as in size) of these images. AEs map different values into different ranges \cite{wang2016auto, chakraborty2019integration}. Slight variations (noise) are treated differently than major variations (change). Thus, it is intuitive that AEs can be used to reduce dimensionality of HSIs because if a pixel has changed its value between two images, the two values will be mapped to different ranges.

CAEs are good for image data as they take both spectral as well as spatial information into consideration. A standard undercomplete convolutional autoencoder would alter the size of the images using pooling and upsampling operation as well as change the number of filters through the convolutional and deconvolutional operations. They are primarily equipped with ReLU (rectified linear units) as activation along the layers. The number of filters in each layer successively increase in the encoder part till the middle layer, and then the number of filters in each layer successively decrease in the decoder part. The proposed FFCAE architecture varies vastly from the standard architecture.

The proposed architecture [Figure \ref{fig:FFCAE}] is different from any traditional convolutional autoencoder architecture. It has exploited the possibility of playing with the size as well as the number of filters in a convolutional autoencoder and added skip connections to create a fusion of lower level and higher level features. In the proposed architecture, the original image is convolved with filters of two different sizes successively. This results in the generation of lower level features with different receptive fields. The features are then concatenated and is convolved with another filter to generate a higher level feature representation. Then both the lower level features and the higher level features are concatenated with a skip connection at the next level (which is the middle code layer). This is the part where feature fusion is taking place. The decoder then convolves the fused feature with filters of two different sizes again and then the concatenated feature maps is convolved with another filter to reconstruct the original image back.

Choosing the window size and the number of filters in a CAE is crucial because the entire physics behind the feature extraction property of a CAE depends on these two parameters mainly. Usually the filter windows in a CAE are taken to be odd i.e. $1 \times 1$, $3 \times 3$, $5 \times 5$, $7 \times 7$, etc depending on the type of image ($1 \times 1$ is usually not taken as it implies no spatial information is considered). The windows $1 \times 1$ and $3 \times 3$ are considered to be smaller size and $5 \times 5$, $7 \times 7$, etc are considered as larger sizes. If the window size is taken smaller, then it may miss some spatial contextual information but as the number of parameters is less, it would be less prone to overfitting. On the other hand, larger window size might cause overfitting but can capture spatial information well. We can combine these two effects by using a multi-window approach during feature extraction. Window size in a CAE is equivalent to the receptive field (i.e. how much of the image the filter sees at a time). Effect of larger receptive field can be achieved by increasing depth. However, going too deep is also problematic for training sometimes. Depth captures higher level features which imply strong object level semantic information and the first layer extracts lower level features which have acute discriminative capabilities. However, there needs to be a balance. Acute discriminative features get influenced by noise while strong object level features often ignore tiny changes. Thus, a fusion of higher level and lower level features may create this balance.

In the proposed architecture, twin filters are used ($3 \times 3$ and $5 \times 5$) i.e. $n_1=3$ and $n_2=5$. Adding another layer extracted higher level features, and inherently increased the receptive field. We chose $n_3=3$ as it gave the effect of a $5 \times 5$ and $7 \times 7$ receptive field. Increasing the depth would have increased the receptive field even more, but that would incorporate additional burden of increased number of parameters. Using two filters ensure that both the spatial information as well as subtle changes are captured. The fusion of lower level features and higher level features is obtained through a skip connection. The code-layer therefore produces the most compact transformed image which would have both the spatial as well as spectral information of the change. It must be noted that even after passing through the layers of the network, the image size is not reduced below its actual size anywhere. This was purposely done to preserve the resolution in the extracted feature maps. After the FFCAE network is trained to reconstruct the input back, the encoder part may be used to extract the deep feature maps (DFM) from the middle code layer.

The proposed FFCAE being an extension of CAEs is structurally built to extract features to reconstruct the input. Since there is no label information provided to the network, it captures some common generic features shared by both the images. However, these features are mapped in different feature maps than the discriminating ones. We can, therefore, easily discard such feature maps which are same for both images using a simple entropy based filtering i.e. reject the feature maps for which,

\begin{equation}
\epsilon(f(I_1) \ominus f(I_2)) = 0,
\end{equation}

where, $\epsilon()$ is the entropy of an image. The feature maps which are selected (Sel-DFM) are used for the further change analysis. The overall procedure of change detection is shown in Figure \ref{fig:overall}.

The trick is that, the same FFCAE was trained with only two images for 50 epochs. It is done with the intuition that most of the pixels in the two images are almost identical as they have no change. So, it would be much more gracious to train a single network to reconstruct the two images.
\section{Results}\label{Results}
The proposed method was implemented and compared with six state-of-the art change detection methods. In order to have a fair comparison, the methods which use any kind of labelling (supervised, semi-supervised, active etc) are excluded. Most of the methods in the literature have tested their methods on either one or two hyperspectral datasets. This is mainly due to the lack of hyperspectral datasets with available ground truth. However, the present manuscript has made an attempt to incorporate as many datasets as possible into the experimentation.
\subsection{Performance Metrics Used}
To quantify the performance of the proposed method, five performance metrics are used, namely, overall accuracy (OA),	Cohen's kappa measure ($\kappa$), $f-score$, percentage of wrong classification (PWC) and Detection Rate (DR). The metrics are calculated from the confusion matrix given by Table \ref{CM}.

\begin{table}[htp]
\centering
\caption{Confusion Matrix}
\begin{tabular}{cl|l|l|}
\cline{3-4}
                             &          & \multicolumn{2}{c|}{Actual}   \\ \cline{3-4}
                             &          & Positive      & Negative      \\
                             &          & Class         & Class         \\ \hline
\multicolumn{1}{|c|}{}                            & Positive & True          & False         \\
\multicolumn{1}{|c|}{Predicted}                    & Class    & Positive (TP) & Positive (FP) \\ \cline{2-4}
\multicolumn{1}{|c|}{}                             & Negative & False         & True          \\
\multicolumn{1}{|c|}{}                             & Class    & Negative (FN) & Negative (TN) \\ \hline
\end{tabular}
\label{CM}
\end{table}

The overall accuracy suggests how accurately the model detects both the changed as well as unchanged pixels.
\begin{equation}
OA=\dfrac{TP+TN}{TP+FP+TN+FN}
\end{equation}
The Cohen's kappa measure takes into account actual agreement and chanced agreement. So, it is a much preferable measure for change detection purposes as the number of changed pixels is often less than the number of unchanged pixels. It is given by,
\begin{equation}
\kappa=\dfrac{OA-CA}{1-CA},
\end{equation}
where,
OA is the overall accuracy or the actual agreement while CA is the accuracy by chance written as,
\begin{equation}
CA=CA_{Class U}+CA_{Class C}.
\end{equation}
Here, the accuracy obtained by chance for unchanged class ($CA_{Class U}$) and changed class ($CA_{Class C}$) are given by,
\begin{equation}
CA_{Class U}=\dfrac{(TP+FP)\times(TP+FN)}{(TP+TN+FP+FN)^2},
\end{equation}
and
\begin{equation}
CA_{Class C}=\dfrac{(TN+FN)\times(TN+FP)}{(TP+TN+FP+FN)^2}.
\end{equation}
The $f-score$ is given by,
\begin{equation}
f-score=\dfrac{2\times precision \times recall}{precision + recall}
\end{equation}
where,
\begin{equation}
precision=\dfrac{TP}{TP + FP},
\end{equation}
and
\begin{equation}
recall=\dfrac{TP}{TP + FN}.
\end{equation}
The percentage of wrong classification is given by,
\begin{equation}
PWC=\dfrac{100 \times (FP+FN)}{TP+FP+TN+FN}\%.
\end{equation}
A lower value of PWC means that the model is better.

FNR (false negative rate) suggests the rate as to how often the method predicts a pixel as changed when actually it is not. It is given by
\begin{equation}
FNR=\dfrac{FN}{FN+TN}.
\end{equation}
TNR (true negative rate) suggests the rate as to how often the method predicts a pixel as a change when actually it is a change. It can be represented as
\begin{equation}
TNR=\dfrac{TN}{FP+TN}.
\end{equation}
Any good change detection method should be able to give high TNR and low FNR. So, we can say, the detection rate (DR) \cite{chakraborty2019integration}, where
\begin{equation}
DR=(1-FNR)\times TNR,
\end{equation}
is high for a better change detection method.
\subsection{Comparison of models}
For comparison, Window-based PCA+kmeans \cite{celik2009unsupervised}, IRMAD \cite{nielsen2007regularized}, SFI-IRMAD \cite{wang2015application}, SFI-DSP \cite{han2017unsupervised} and DSFANet \cite{du2019unsupervised} are used. S3DCAE \cite{zhao2021spectral} is used for comparison as the deep features extracted by it could be used for unsupervised change detection too. All these methods have used K-means as the segmentation or clustering algorithm for differentiating the changed regions from the unchanged regions. So, the proposed method has also been analysed with K-means as the function $g()$. The difference operator $\ominus$ has been chosen to be absolute difference (AD) between the feature maps and spectral angle mapper (SAM) \cite{carvalho2011new} as the comparison methods have used either one of the two.

\begin{figure}
\centering
\begin{subfigure}{0.3\textwidth}
\centering
\includegraphics[width=0.4\linewidth]{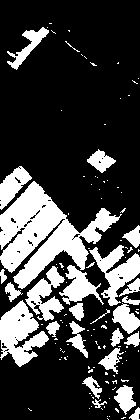}
\caption{}
\end{subfigure}
\begin{subfigure}{0.3\textwidth}
\centering
\includegraphics[width=0.4\linewidth]{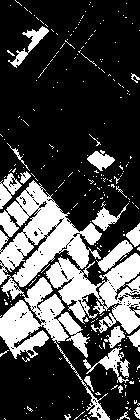}
\caption{}
\end{subfigure}
\begin{subfigure}{0.3\textwidth}
\centering
\includegraphics[width=0.4\linewidth]{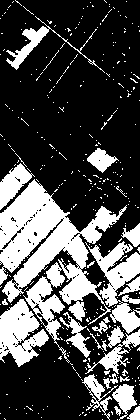}
\caption{}
\end{subfigure}
\begin{subfigure}{0.3\textwidth}
\centering
\includegraphics[width=0.4\linewidth]{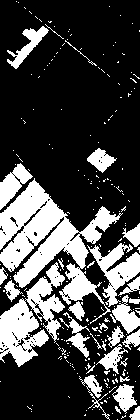}
\caption{}
\end{subfigure}
\begin{subfigure}{0.3\textwidth}
\centering
\includegraphics[width=0.4\linewidth]{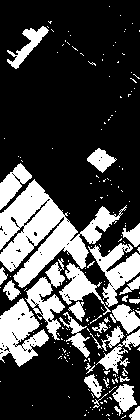}
\caption{}
\end{subfigure}
\begin{subfigure}{0.3\textwidth}
\centering
\includegraphics[width=0.4\linewidth]{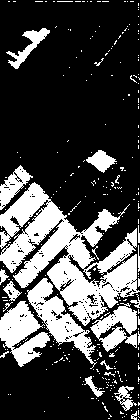}
\caption{}
\end{subfigure}
\begin{subfigure}{0.3\textwidth}
\centering
\includegraphics[width=0.4\linewidth]{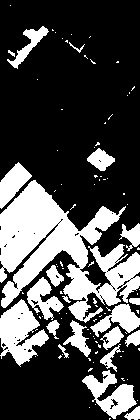}
\caption{}
\end{subfigure}
\begin{subfigure}{0.3\textwidth}
\centering
\includegraphics[width=0.4\linewidth]{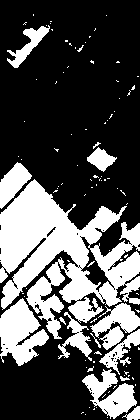}
\caption{}
\end{subfigure}
\begin{subfigure}{0.3\textwidth}
\centering
\includegraphics[width=0.4\linewidth]{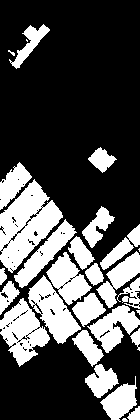}
\caption{}
\end{subfigure}
\caption{Comparison of results for China Dataset: (a) Windows PCA, (b) IRMAD, (c) SFI-IRMAD, (d) SFI-DSP, (e) S3DCAE-AD (f) Deep SFA, (g) FFCAE-SAM, (h) FFCAE-AD, (i) Ground Truth}
\label{fig:figChina}
\end{figure}

\begin{figure}
\centering
\begin{subfigure}{0.3\textwidth}
\centering
\includegraphics[width=1\linewidth]{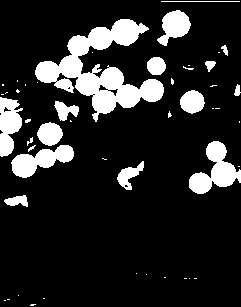}
\caption{}
\end{subfigure}
\begin{subfigure}{0.3\textwidth}
\centering
\includegraphics[width=1\linewidth]{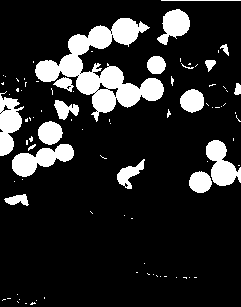}
\caption{}
\end{subfigure}
\begin{subfigure}{0.3\textwidth}
\centering
\includegraphics[width=1\linewidth]{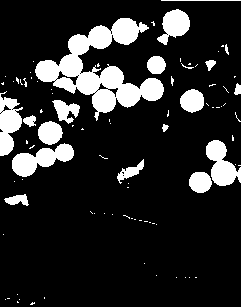}
\caption{}
\end{subfigure}
\begin{subfigure}{0.3\textwidth}
\centering
\includegraphics[width=1\linewidth]{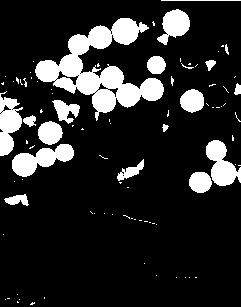}
\caption{}
\end{subfigure}
\begin{subfigure}{0.3\textwidth}
\centering
\includegraphics[width=1\linewidth]{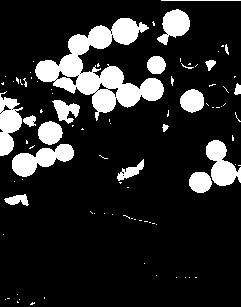}
\caption{}
\end{subfigure}
\begin{subfigure}{0.3\textwidth}
\centering
\includegraphics[width=1\linewidth]{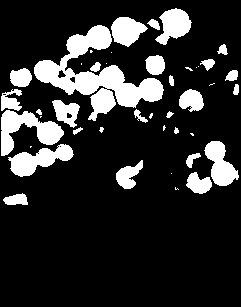}
\caption{}
\end{subfigure}
\begin{subfigure}{0.3\textwidth}
\centering
\includegraphics[width=1\linewidth]{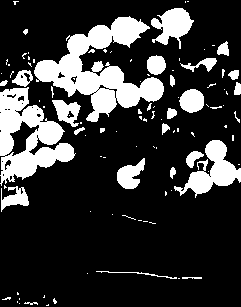}
\caption{}
\end{subfigure}
\begin{subfigure}{0.3\textwidth}
\centering
\includegraphics[width=1\linewidth]{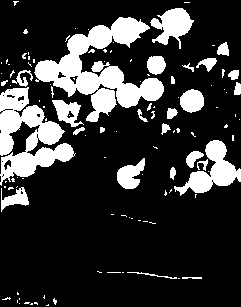}
\caption{}
\end{subfigure}
\begin{subfigure}{0.3\textwidth}
\centering
\includegraphics[width=1\linewidth]{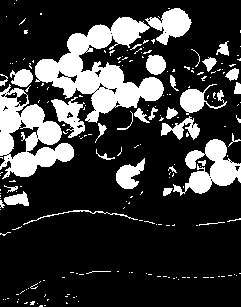}
\caption{}
\end{subfigure}
\caption{Comparison of results for USA Dataset: (a) Windows PCA, (b) IRMAD, (c) SFI-IRMAD, (d) SFI-DSP, (e) S3DCAE-AD (f) Deep SFA, (g) FFCAE-SAM, (h) FFCAE-AD, (i) Ground Truth}
\label{fig:figHermiston}
\end{figure}

\begin{figure}
\centering
\begin{subfigure}{0.3\textwidth}
\centering
\includegraphics[width=0.6\linewidth]{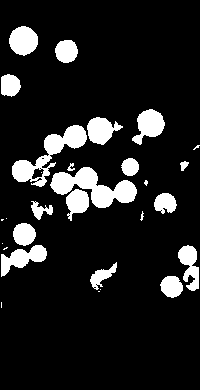}
\caption{}
\end{subfigure}
\begin{subfigure}{0.3\textwidth}
\centering
\includegraphics[width=0.6\linewidth]{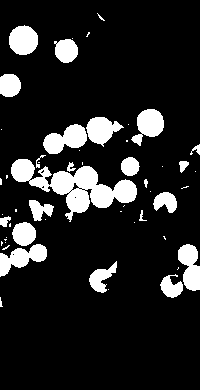}
\caption{}
\end{subfigure}
\begin{subfigure}{0.3\textwidth}
\centering
\includegraphics[width=0.6\linewidth]{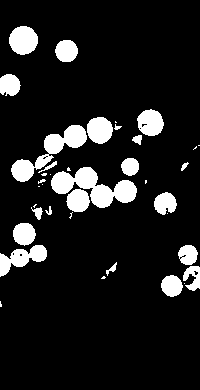}
\caption{}
\end{subfigure}
\begin{subfigure}{0.3\textwidth}
\centering
\includegraphics[width=0.6\linewidth]{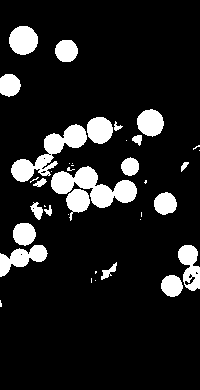}
\caption{}
\end{subfigure}
\begin{subfigure}{0.3\textwidth}
\centering
\includegraphics[width=0.6\linewidth]{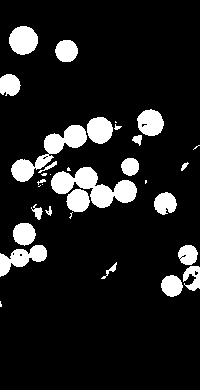}
\caption{}
\end{subfigure}
\begin{subfigure}{0.3\textwidth}
\centering
\includegraphics[width=0.6\linewidth]{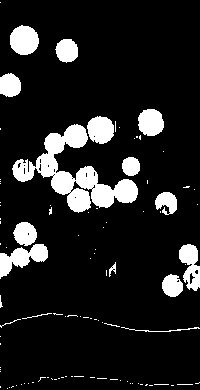}
\caption{}
\end{subfigure}
\begin{subfigure}{0.3\textwidth}
\centering
\includegraphics[width=0.6\linewidth]{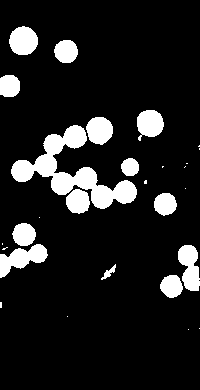}
\caption{}
\end{subfigure}
\begin{subfigure}{0.3\textwidth}
\centering
\includegraphics[width=0.6\linewidth]{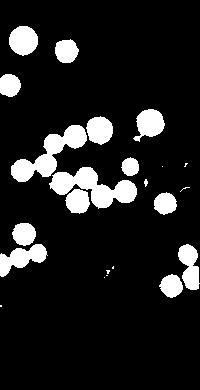}
\caption{}
\end{subfigure}
\begin{subfigure}{0.3\textwidth}
\centering
\includegraphics[width=0.6\linewidth]{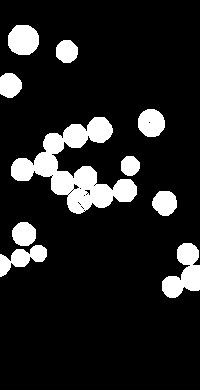}
\caption{}
\end{subfigure}
\caption{Comparison of results for River Dataset: (a) Windows PCA, (b) IRMAD, (c) SFI-IRMAD, (d) SFI-DSP, (e) S3DCAE-AD (f) Deep SFA, (g) FFCAE-SAM, (h) FFCAE-AD, (i) Ground Truth}
\label{fig:figRiver}
\end{figure}

\begin{figure}
\centering
\begin{subfigure}{0.3\textwidth}
\centering
\includegraphics[width=0.7\linewidth]{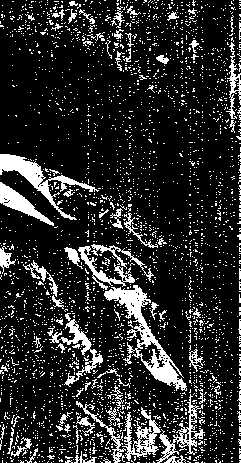}
\caption{}
\end{subfigure}
\begin{subfigure}{0.3\textwidth}
\centering
\includegraphics[width=0.7\linewidth]{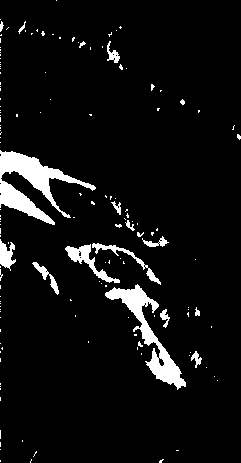}
\caption{}
\end{subfigure}
\begin{subfigure}{0.3\textwidth}
\centering
\includegraphics[width=0.7\linewidth]{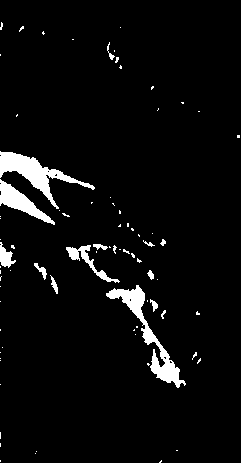}
\caption{}
\end{subfigure}
\begin{subfigure}{0.3\textwidth}
\centering
\includegraphics[width=0.7\linewidth]{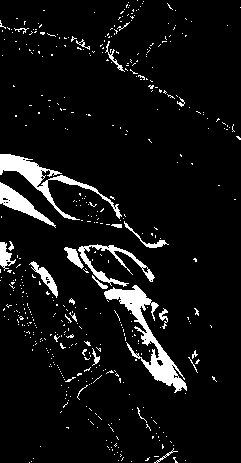}
\caption{}
\end{subfigure}
\begin{subfigure}{0.3\textwidth}
\centering
\includegraphics[width=0.7\linewidth]{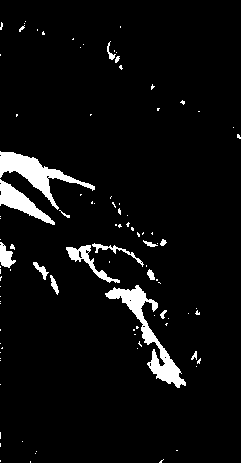}
\caption{}
\end{subfigure}
\begin{subfigure}{0.3\textwidth}
\centering
\includegraphics[width=0.7\linewidth]{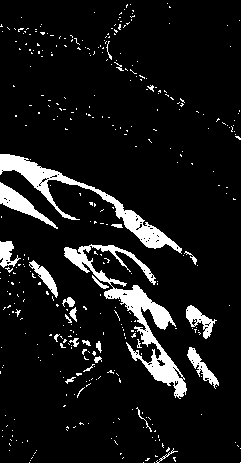}
\caption{}
\end{subfigure}
\begin{subfigure}{0.3\textwidth}
\centering
\includegraphics[width=0.7\linewidth]{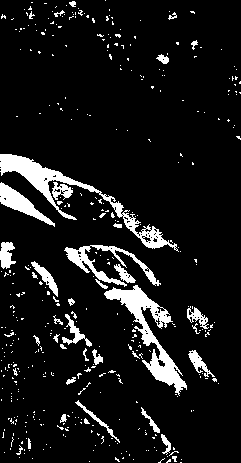}
\caption{}
\end{subfigure}
\begin{subfigure}{0.3\textwidth}
\centering
\includegraphics[width=0.7\linewidth]{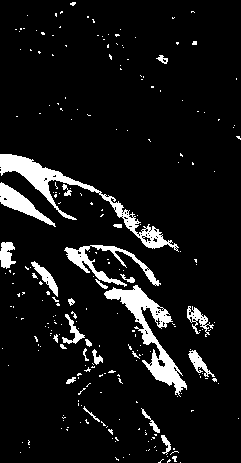}
\caption{}
\end{subfigure}
\begin{subfigure}{0.3\textwidth}
\centering
\includegraphics[width=0.7\linewidth]{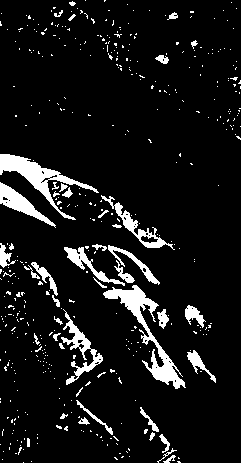}
\caption{}
\end{subfigure}
\caption{Comparison of results for Hermiston Dataset: ((a) Windows PCA, (b) IRMAD, (c) SFI-IRMAD, (d) SFI-DSP, (e) S3DCAE-AD (f) Deep SFA, (g) FFCAE-SAM, (h) FFCAE-AD, (i) Ground Truth}
\label{fig:figUSA}
\end{figure}

The results for the image pair datasets are given in Tables \ref{tab:china_results}-\ref{tab:Hermiston_results}. The results marked in bold represent the best performance and those marked in red represent the second best performance on the particular dataset. Although, it may seem a bit misleading to use accuracy for an unsupervised method, but since we already have the ground truth to compare with, we can actually see how accurately our proposed method can identify the actual changes. The detected change map for the various models are reported in Figures \ref{fig:figChina}-\ref{fig:figUSA}.

\begin{table}[htp]
\centering
\caption{Results for China Dataset}
\begin{tabular}{llllll}
\hline
 & OA & $\kappa$ & $f-score$ & PWC & DR \\ \hline\hline
Windows & 0.8688 & 0.6801 & 0.8429 & 13.1241 & 0.7673 \\
PCA &  &  &  &  &  \\ \hline
IRMAD & 0.8659 & 0.6801 & 0.8408 & 13.4150 & 0.7864 \\ \hline
SFI & 0.8784 & 0.7165 & 0.8583 & 12.1582 & 0.8284 \\
IRMAD &  &  &  &  &  \\ \hline
SFI-DSP & 0.8817 & 0.7168 & 0.8595 & 11.8316 & 0.8020 \\ \hline
S3DCAE-AD & 0.8885 & 0.7333 & 0.8677 & 11.1463 & 0.8108 \\ \hline
Deep SFA & 0.8920 & 0.7419 & 0.8719 & 10.8010 & 0.8166 \\ \hline
FFCAE-SAM & \textcolor{red}{0.8934} & \textcolor{red}{0.7489} & \textcolor{red}{0.8746} & \textcolor{red}{10.6650} & \textcolor{red}{0.8360} \\ \hline
FFCAE-AD & \textbf{0.8956} & \textbf{0.7604} & \textbf{0.8804} & \textbf{10.4422} & \textbf{0.8725} \\ \hline
\end{tabular}
\label{tab:china_results}
\end{table}

\begin{table}[htp]
\centering
\caption{Results for USA Dataset}
\begin{tabular}{llllll}
\hline
 & OA & $\kappa$ & $f-score$ & PWC & DR \\ \hline\hline
Windows & 0.9106 & 0.7111 & 0.8662 & 8.9435 & 0.8182 \\
PCA &  &  &  &  &  \\ \hline
IRMAD & 0.9217 & 0.7471 & 0.8847 & 7.8298 & 0.8303 \\ \hline
SFI & 0.9230 & 0.7518 & 0.8868 & 7.6959 & 0.8325 \\
IRMAD &  &  &  &  &  \\ \hline
SFI-DSP & 0.9228 & 0.7508 & 0.8863 & 7.7230 & 0.8319 \\ \hline
S3DCAE-AD & 0.9229 & 0.7511 & 0.8865 & 7.7135 & 0.8321 \\ \hline
Deep SFA & 0.9317 & 0.7867 & 0.8990 & 6.8323 & 0.8592 \\ \hline
FFCAE-SAM & \textbf{0.9462} & \textbf{0.8421} & \textbf{0.9215} & \textbf{5.3780} & \textbf{0.9143} \\ \hline
FFCAE-AD & \textcolor{red}{0.9447} & \textcolor{red}{0.8360} & \textcolor{red}{0.9189} & \textcolor{red}{5.5321} & \textcolor{red}{0.9057} \\ \hline
\end{tabular}
\label{tab:USA_results}
\end{table}

\begin{table}[htp]
\centering
\caption{Results for River Dataset}
\begin{tabular}{llllll}
\hline
 & OA & $\kappa$ & $f-score$ & PWC & DR \\ \hline\hline
Windows & 0.8863 & 0.4879 & 0.7630 & 11.3736 & \textcolor{red}{0.9562} \\
PCA &  &  &  &  &  \\ \hline
IRMAD & 0.9401 & 0.5663 & 0.7888 & 5.9920 & 0.9116 \\ \hline
SFI & 0.9415 & 0.5311 & 0.7846 & 5.8530 & 0.8984 \\
IRMAD &  &  &  &  &  \\ \hline
SFI-DSP & 0.9491 & 0.6464 & 0.8261 & 5.0886 & 0.9274 \\ \hline
S3DCAE-AD & 0.9420 & 0.5535 & 0.7896 & 5.7966 & 0.9044 \\ \hline
Deep SFA & 0.9461 & 0.6645 & 0.8323 & 5.3933 & 0.9443 \\ \hline
FFCAE-SAM & \textcolor{red}{0.9584} & \textbf{0.7454} & \textbf{0.8730} & \textcolor{red}{4.1646} & \textbf{0.9610} \\ \hline
FFCAE-AD & \textbf{0.9604} & \textcolor{red}{0.7436} & \textcolor{red}{0.8720} & \textbf{3.9630} & 0.9519 \\ \hline
\end{tabular}
\label{tab:River_results}
\end{table}

\begin{table}[htp]
\centering
\caption{Results for Hermiston Dataset}
\begin{tabular}{llllll}
\hline
 & OA & $\kappa$ & $f-score$ & PWC & DR \\ \hline\hline
Windows & 0.9670 & 0.8503 & 0.9252 & 3.3013 & 0.9584 \\
PCA &  &  &  &  &  \\ \hline
IRMAD & 0.9618 & 0.8348 & 0.9178 & 3.8154 & 0.9679 \\ \hline
SFI & 0.9751 & 0.8837 & 0.9426 & 2.4936 & 0.9588 \\
IRMAD &  &  &  &  &  \\ \hline
SFI-DSP & 0.9752 & 0.8872 & 0.9437 & 2.4795 & 0.9666 \\ \hline
S3DCAE-AD & 0.9751 & 0.8842 & 0.9427 & 2.4872 & 0.9594 \\ \hline
Deep SFA & 0.9742 & 0.8833 & 0.9417 & 2.5795 & 0.9674 \\ \hline
FFCAE-SAM & \textcolor{red}{0.9812} & \textcolor{red}{0.9145} & \textcolor{red}{ 0.9574} & \textcolor{red}{1.8756} & \textcolor{red}{0.9727} \\ \hline
FFCAE-AD & \textbf{0.9819} & \textbf{0.9176} & \textbf{0.9589} & \textbf{1.8051} & \textbf{0.9730} \\ \hline
\end{tabular}
\label{tab:Hermiston_results}
\end{table}


\subsection{Statistical-significance test}
The performance of different models is assessed by several performance scores. We should not, however, accept any of the metrics in isolation to prove that one is better than the other. No metric alone can encompass every area of interest. Even for one element of interest, several metrics must be given to detail the performance of a model. The statistical tests often evaluate the superiority of the a model by encompassing a single metric only. Therefore, this manuscript shows a ranked ordering of the models taking all the performance metrics into account. The average ranking of the compared models on all the four datasets on the five metrics are reported in the Table \ref{tab:Ranking}. To avoid rewriting the names of the models, the models are denoted as A (Windows PCA), B (IRMAD), C (SFI-IRMAD), D (SFI-DSP), E (S3DCAE-AD), F (Deep SFA), G (FFCAE-SAM) and H (FFCAE-AD).
\begin{table}[tbp]
\caption{Average ranking of the compared models on all the four datasets on the five metrics.}
\centering
\begin{tabular}{lllllllll}
 & A & B & C & D & E & F & G & H \\\hline\hline
Accuracy & 7.50 & 7.50 & 5.25 & 4.25 & 4.50 & 4.00 & 1.75 & 1.25 \\
Kappa & 7.75 & 6.75 & 5.50 & 4.50 & 4.75 & 3.75 & 1.50 & 1.50 \\
f-score & 7.50 & 7.25 & 5.50 & 4.50 & 4.50 & 3.75 & 1.50 & 1.50 \\
PWC & 7.50 & 7.50 & 5.25 & 4.25 & 4.50 & 4.00 & 1.75 & 1.25 \\
DR & 6.50 & 5.75 & 5.50 & 5.50 & 5.75 & 3.75 & 1.50 & 1.75 \\\hline
\end{tabular}
\label{tab:Ranking}
\end{table}
\begin{table}[tbp]
\caption{Value of Tukey HSD Q-statistic for the models compared.}
\centering
\begin{tabular}{lllllllll}
 & A & B & C & D & E & F & G & H \\\hline\hline
A & 0 & \textcolor{red}{2.11} & \textbf{10.26} & \textbf{14.47} & \textbf{13.42} & \textbf{18.43} & \textbf{30.27} & \textbf{31.06} \\
B &  & 0 & \textbf{8.16} & \textbf{12.37} & \textbf{11.32} & \textbf{16.32} & \textbf{28.16} & \textbf{28.95} \\
C &  &  & 0 & \textcolor{red}{4.21} & \textcolor{red}{3.16} & \textbf{8.16} & \textbf{20.00} & \textbf{20.79} \\
D &  &  &  & 0 & \textcolor{red}{1.05} & \textcolor{red}{3.95} & \textbf{15.79} & \textbf{16.58} \\
E &  &  &  &  & 0 & \textbf{5.00} & \textbf{16.84} & \textbf{17.63} \\
F &  &  &  &  &  & 0 & \textbf{11.84} & \textbf{12.63} \\
G &  &  &  &  &  &  & 0 & \textcolor{red}{0.79} \\
H &  &  &  &  &  &  &  & 0\\\hline
\end{tabular}
\label{tab:Tukey}
\end{table}
It can be seen that the proposed FFCAE with SAM and AD have more or less similar ranks, but are better in ranks than the others. For statistical test, one can perform a Tukey HSD test \cite{abdi2010tukey} as a post-hoc test to justify the superiority of the proposed model. There are eight models ($k=8$) being compared against each other and thus 28 pairs of models to compare. Degrees of freedom for the error term is $\nu=k(k+1)=32$. The error obtained is 5.775 and since the degree of freedom for error $\nu=32$, the mean-square error, $MSE=5.775/\nu=0.1805$. Thus, the Tukey HSD Q-statistic is given as,

\begin{equation}
Q_{i,j}=|\bar{r}_i-\bar{r}_j|.\sqrt{\dfrac{n}{MSE}}
\end{equation}

The critical value at 95\% confidence ($\alpha=0.05$ significance level) $Q_{critical}^{\alpha=0.05, k=8, \nu=32}=4.5209$. Table \ref{tab:Tukey} shows the Tukey HSD Q-statistic for the compared models. Any value greater than $Q_{critical}^{\alpha=0.05, k=8, \nu=32}$ implies significance. The values marked in bold suggest that the difference is significant and those marked in red suggest insignificance. It is evident that the proposed models FFCAE-SAM and FFCAE-AD (renamed as G and H, respectively for simplicity) are significantly better than all the other models. However, there is no significant difference between FFCAE-SAM and FFCAE-AD which is quite intuitive as the proposed feature extraction method is the same for both.
\section{Conclusion and Future Work}\label{conclusion}
We can safely claim that the proposed method of feature extraction performed better than all the methods compared with in all the four datasets and all the five metrics. The proposed method takes only twin filters into account. Increasing the number of filters would give rise to situations of overfitting and using filters of size more than $5 \times 5$ would only contribute to the effect of suppressing the small changes. However, the consequence of increasing the depth in detecting changes is a matter of further investigation and needs a separate study in itself. In that case, the presence or absence of skip connections may be further investigated from the direction of residual networks.

This manuscript also has not explored the capabilities of FFCAE feature extraction in a supervised or semi-supervised setting and concerns only with unsupervised binary change detection. We are planning to explore this model and possibly adapt it to be applicable to other supervised or semi-supervised classification tasks concerning such high dimensional remotely sensed images in general. Including the label information in the dimensionality reduction process may also be extended to multi-class change detection as well.

\bibliographystyle{IEEEtran}
\bibliography{Bibliography}

\begin{thebibliography}{10}
\providecommand{\url}[1]{#1}
\csname url@samestyle\endcsname
\providecommand{\newblock}{\relax}
\providecommand{\bibinfo}[2]{#2}
\providecommand{\BIBentrySTDinterwordspacing}{\spaceskip=0pt\relax}
\providecommand{\BIBentryALTinterwordstretchfactor}{4}
\providecommand{\BIBentryALTinterwordspacing}{\spaceskip=\fontdimen2\font plus
\BIBentryALTinterwordstretchfactor\fontdimen3\font minus
  \fontdimen4\font\relax}
\providecommand{\BIBforeignlanguage}[2]{{%
\expandafter\ifx\csname l@#1\endcsname\relax
\typeout{** WARNING: IEEEtran.bst: No hyphenation pattern has been}%
\typeout{** loaded for the language `#1'. Using the pattern for}%
\typeout{** the default language instead.}%
\else
\language=\csname l@#1\endcsname
\fi
#2}}
\providecommand{\BIBdecl}{\relax}
\BIBdecl

\bibitem{jin2018intrinsic}
X.~Jin, Y.~Gu, and T.~Liu, ``Intrinsic {I}mage {R}ecovery from {R}emote
  {S}ensing {H}yperspectral {I}mages,'' \emph{IEEE Transactions on Geoscience
  and Remote Sensing}, vol.~57, no.~1, pp. 224--238, 2018.

\bibitem{volpi2013supervised}
M.~Volpi, D.~Tuia, F.~Bovolo, M.~Kanevski, and L.~Bruzzone, ``Supervised
  {C}hange {D}etection in {V}{H}{R} {I}mages using {C}ontextual {I}nformation
  and {S}upport {V}ector {M}achines,'' \emph{International Journal of Applied
  Earth Observation and Geoinformation}, vol.~20, pp. 77--85, 2013.

\bibitem{wu2017semi}
H.~Wu and S.~Prasad, ``Semi-supervised {D}eep {L}earning using {P}seudo
  {L}abels for {H}yperspectral {I}mage {C}lassification,'' \emph{IEEE
  Transactions on Image Processing}, vol.~27, no.~3, pp. 1259--1270, 2017.

\bibitem{yuan2015semi}
Y.~Yuan, H.~Lv, and X.~Lu, ``Semi-supervised {C}hange {D}etection {M}ethod for
  {M}ulti-temporal {H}yperspectral {I}mages,'' \emph{Neurocomputing}, vol. 148,
  pp. 363--375, 2015.

\bibitem{liu2014hierarchical}
S.~Liu, L.~Bruzzone, F.~Bovolo, and P.~Du, ``Hierarchical {U}nsupervised
  {C}hange {D}etection in {M}ultitemporal {H}yperspectral {I}mages,''
  \emph{IEEE Transactions on Geoscience and Remote Sensing}, vol.~53, no.~1,
  pp. 244--260, 2014.

\bibitem{du2019unsupervised}
B.~Du, L.~Ru, C.~Wu, and L.~Zhang, ``Unsupervised {D}eep {S}low {F}eature
  {A}nalysis for {C}hange {D}etection in {M}ulti-temporal {R}emote {S}ensing
  {I}mages,'' \emph{IEEE Transactions on Geoscience and Remote Sensing},
  vol.~57, no.~12, pp. 9976--9992, 2019.

\bibitem{wang2018getnet}
Q.~Wang, Z.~Yuan, Q.~Du, and X.~Li, ``G{E}{T}{N}{E}{T}: {A} {G}eneral
  {E}nd-to-end 2-d {C}{N}{N} {F}ramework for {H}yperspectral {I}mage {C}hange
  {D}etection,'' \emph{IEEE Transactions on Geoscience and Remote Sensing},
  vol.~57, no.~1, pp. 3--13, 2018.

\bibitem{zhan2017change}
Y.~Zhan, K.~Fu, M.~Yan, X.~Sun, H.~Wang, and X.~Qiu, ``Change {D}etection based
  on {D}eep {S}iamese {C}onvolutional {N}etwork for {O}ptical {A}erial
  {I}mages,'' \emph{IEEE Geoscience and Remote Sensing Letters}, vol.~14,
  no.~10, pp. 1845--1849, 2017.

\bibitem{zhao2021spectral}
C.~Zhao, H.~Cheng, and S.~Feng, ``A {S}pectral-{S}patial {C}hange {D}etection
  {M}ethod {B}ased on {S}implified 3-{D} {C}onvolutional {A}utoencoder for
  {M}ultitemporal {H}yperspectral {I}mages,'' \emph{IEEE Geoscience and Remote
  Sensing Letters}, pp. 1--5, 2021.

\bibitem{de2013semi}
F.~De~Morsier, D.~Tuia, M.~Borgeaud, V.~Gass, and J.~P. Thiran,
  ``Semi-supervised {N}ovelty {D}etection using {S}{V}{M} {E}ntire {S}olution
  {P}ath,'' \emph{IEEE Transactions on Geoscience and Remote Sensing}, vol.~51,
  no.~4, pp. 1939--1950, 2013.

\bibitem{roy2013neural}
M.~Roy, S.~Ghosh, and A.~Ghosh, ``A {N}eural {A}pproach under {A}ctive
  {L}earning {M}ode for {C}hange {D}etection in {R}emotely {S}ensed {I}mages,''
  \emph{IEEE Journal of Selected Topics in Applied Earth Observations and
  Remote Sensing}, vol.~7, no.~4, pp. 1200--1206, 2013.

\bibitem{ghosh2007context}
S.~Ghosh, L.~Bruzzone, S.~Patra, F.~Bovolo, and A.~Ghosh, ``A
  {C}ontext-{S}ensitive {T}echnique for {U}nsupervised {C}hange {D}etection
  based on {H}opfield-type {N}eural {N}etworks,'' \emph{IEEE Transactions on
  Geoscience and Remote Sensing}, vol.~45, no.~3, pp. 778--789, 2007.

\bibitem{celik2009unsupervised}
T.~Celik, ``Unsupervised {C}hange {D}etection in {S}atellite {I}mages using
  {P}rincipal {C}omponent {A}nalysis and $ k $-means {C}lustering,'' \emph{IEEE
  Geoscience and Remote Sensing Letters}, vol.~6, no.~4, pp. 772--776, 2009.

\bibitem{nielsen2007regularized}
A.~A. Nielsen, ``The {R}egularized {I}teratively {R}eweighted {M}{A}{D}
  {M}ethod for {C}hange {D}etection in {M}ulti-and {H}yperspectral {D}ata,''
  \emph{IEEE Transactions on Image processing}, vol.~16, no.~2, pp. 463--478,
  2007.

\bibitem{wang2015application}
B.~Wang, S.~K. Choi, Y.~K. Han, S.~K. Lee, and J.~W. Choi, ``Application of
  {I}{R}-{M}{A}{D} using {S}ynthetically {F}used {I}mages for {C}hange
  {D}etection in {H}yperspectral {D}ata,'' \emph{Remote Sensing Letters},
  vol.~6, no.~8, pp. 578--586, 2015.

\bibitem{han2017unsupervised}
Y.~K. Han, A.~Chang, S.~K. Choi, H.~Park, and J.~Choi, ``An {U}nsupervised
  {A}lgorithm for {C}hange {D}etection in {H}yperspectral {R}emote {S}ensing
  {D}ata using {S}ynthetically {F}used {I}mages and {D}erivative {S}pectral
  {P}rofiles,'' \emph{Journal of Sensors}, vol. 2017, 2017.

\bibitem{carvalho2011new}
O.~A. Carvalho~J{\'u}nior, R.~F. Guimar{\~a}es, A.~R. Gillespie, N.~C. Silva,
  and R.~A.~T. Gomes, ``A {N}ew {A}pproach to {C}hange {V}ector {A}nalysis
  using {D}istance and {S}imilarity {M}easures,'' \emph{Remote Sensing},
  vol.~3, no.~11, pp. 2473--2493, 2011.

\bibitem{tommaselli2019refining}
A.~M.~G. Tommaselli, L.~D. Santos, R.~A. de~Oliveira, A.~Berveglieri, N.~N.
  Imai, and E.~Honkavaara, ``Refining the {I}nterior {O}rientation of a
  {H}yperspectral {F}rame {C}amera with {P}reliminary {B}ands
  {C}o-{R}egistration,'' \emph{IEEE Journal of Selected Topics in Applied Earth
  Observations and Remote Sensing}, vol.~12, no.~7, pp. 2097--2106, 2019.

\bibitem{hasanlou2018hyperspectral}
M.~Hasanlou and S.~T. Seydi, ``Hyperspectral {C}hange {D}etection: {A}n
  {E}xperimental {C}omparative {S}tudy,'' \emph{International Journal of Remote
  Sensing}, vol.~39, no.~20, pp. 7029--7083, 2018.

\bibitem{javierlopezfandino2018stacked}
\BIBentryALTinterwordspacing
J.~L. Fandino, A.~S. Garea, D.~B. Heras, and F.~Arg\"{u}ello, ``Stacked
  {A}utoencoders for {M}ulticlass {C}hange {D}etection in {H}yperspectral
  {I}mages,'' in \emph{International Geoscience and Remote Sensing Symposium
  IGARSS 2018}.\hskip 1em plus 0.5em minus 0.4em\relax IEEE, 2018, pp.
  1906--1909. [Online]. Available:
  \url{http://dx.doi.org/10.1109/IGARSS.2018.8518338}
\BIBentrySTDinterwordspacing

\bibitem{wang2016auto}
Y.~Wang, H.~Yao, and S.~Zhao, ``Auto-encoder based {D}imensionality
  {R}eduction,'' \emph{Neurocomputing}, vol. 184, pp. 232--242, 2016.

\bibitem{chakraborty2019integration}
D.~Chakraborty, V.~Narayanan, and A.~Ghosh, ``Integration of {D}eep {F}eature
  {E}xtraction and {E}nsemble {L}earning for {O}utlier {D}etection,''
  \emph{Pattern Recognition}, vol.~89, pp. 161--171, 2019.

\bibitem{abdi2010tukey}
H.~Abdi and L.~J. Williams, ``Tukey's {H}onestly {S}ignificant {D}ifference
  ({H}{S}{D}) {T}est,'' \emph{Encyclopedia of Research Design}, vol.~3, no.~1,
  pp. 1--5, 2010.

\end{thebibliography}
\end{document}